# Analysis and Prediction of Pedestrian Crosswalk Behavior during Automated Vehicle Interactions

Suresh Kumaar Jayaraman[1], Dawn M. Tilbury[1], X. Jessie Yang[2], Anuj K. Pradhan[3], and Lionel P. Robert Jr.[4]

*Abstract*— For safe navigation around pedestrians, automated vehicles (AVs) need to plan their motion by accurately predicting pedestrians' trajectories over long time horizons. Current approaches to AV motion planning around crosswalks predict only for short time horizons (1-2 s) and are based on data from pedestrian interactions with human-driven vehicles (HDVs). In this paper, we develop a hybrid systems model that uses pedestrians' gap acceptance behavior and constant velocity dynamics for long-term pedestrian trajectory prediction when interacting with AVs. Results demonstrate the applicability of the model for long-term (> 5 s) pedestrian trajectory prediction at crosswalks. Further, we compared measures of pedestrian crossing behaviors in the immersive virtual environment (when interacting with AVs) to that in the real world (results of published studies of pedestrians interacting with HDVs), and found similarities between the two. These similarities demonstrate the applicability of the hybrid model of AV interactions developed from an immersive virtual environment (IVE) for real-world scenarios for both AVs and HDVs.

## I. INTRODUCTION

A significant challenge for automated vehicles (AVs) is safe interaction with pedestrians, especially at uncontrolled mid-block crosswalks [1], [2]. Thus, it is critical that AVs can reliably predict pedestrian trajectories for safe motion planning [3]. Short-term trajectory predictions [4], [5] may be sufficient for collision avoidance at low vehicle speeds, but at higher speeds, the AVs must be able to predict pedestrian trajectories over long durations (5-10 s) [6].

In this paper, we focus on pedestrians intending to cross. Generally around crosswalks, pedestrians can be doing one of three actions: (1) approach crosswalk, (2) wait, or (3) cross [7]. Additionally, they can (4) walk away from crosswalk. We thereby define *pedestrian crossing behavior* as the sequence of actions and transitions between these actions together with the resulting position trajectory. Individual *measures of crossing behavior* describe pedestrian crossing behavior during the actions or transitions such as 'walking speed' during crossing or 'gap acceptance' during the wait to cross transition.

We develop a novel hybrid systems model that can effectively capture long-term pedestrian crossing behavior and use it to predict long-term pedestrian crossing trajectories. We define a finite state machine, with the four actions as the four discrete states, and with a set of possible transitions between them triggered by pedestrian position or a pedestrian decision (i.e., to wait or to cross). Each discrete state has a continuous dynamics model that defines the evolution of pedestrian position and velocity in that state, thus creating a hybrid systems model. The combination of discrete actions with continuous motion enables long-term prediction of pedestrian crossing behavior. The hybrid systems model was demonstrated to more accurately predict pedestrian crossing trajectories by leveraging standard models such as Support Vector Machine (SVM) gap acceptance that predicts the wait to cross transition, and constant velocity motion models that define the evolution of pedestrian position and velocity. In this paper, we assume the following. Pedestrians are either walking or standing still. If they are walking, they can be crossing, approaching or walking away from the crosswalk. They always walk only on the sidewalk, or across the crosswalk, at a constant velocity. If they are standing still near the crosswalk, they are waiting to cross.

[1]Department of Mechanical Engineering, University of Michigan, Ann Arbor, USA. E-mail: jskumaar, tilbury@umich.edu
[2]Department of Industrial & Operations Engineering, University of Michigan, Ann Arbor, USA. E-mail: xijyang@umich.edu
[3]Department of Mechanical & Industrial Engineering, University of Massachusetts, Amherst, USA. E-mail: anujkpradhan@umass.edu
[4]*Corresponding author*. School of Information, University of Michigan, Ann Arbor, USA. E-mail: lprobert@umich.edu

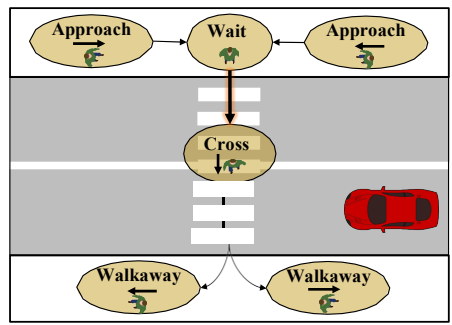

Fig. 1. Pedestrian behaviors when they intend to cross the road. Pedestrians are assumed to use the crosswalk for crossing and at any given time can be doing one of four actions – approaching the crosswalk, waiting near the crosswalk (and deciding when to cross), crossing, or walking away from the crosswalk. The bubbles represent the actions and the arrows represent the action transitions. The bold arrow represents the transition from wait to cross, i.e. the pedestrian's decision to cross.

Current AV planning models [4], [8] use data sets of pedestrian interactions with human-driven vehicles (HDVs) as real-world interactions with AVs are limited. Unlike HDVs, AVs are expected to always stop for pedestrians [9]. The differences in expectations can cause different pedestrian behavior towards AVs. However, existing studies have mixed results on the similarity between pedestrian behavior towards AVs and HDVs [10]–[12], which requires further exploration.

We collected data of pedestrian-AV interactions through a user study in an immersive virtual environment (IVE). Prior works have demonstrated that measures of pedestrian crossing behaviors such as intent to cross, perceived safety of crossing, and walking speeds were similar between IVE and the real world [13], [14]. Building upon this, we compared two measures of crossing behavior—accepted time gaps and walking speeds—between AV interactions in IVE and published results from real-world HDV interactions.

The contributions of this paper are two-fold. First, we contributed to the literature on long-term (> 5 s) pedestrian crossing behavior (trajectory) prediction by developing a hybrid systems model that predicts both high-level action transitions and low-level continuous motion evolution. Second, we demonstrated the applicability of the hybrid model developed from IVE for real-world scenarios with both AVs and HDVs. The rest of this paper is organized as follows. Section II summarizes existing work in pedestrian behavior prediction and pedestrian-AV interactions. Sections III and IV explain our data collection method and proposed hybrid model respectively. Section V reports the performance of our hybrid model compared to a baseline model and compares two measures of pedestrian crossing behavior. Section VI discusses the implications of the study, followed by conclusions and future work in Section VII.

## II. BACKGROUND AND RELATED WORK

### A. Pedestrian Crossing Decision Models

Pedestrian's decision to cross is commonly modeled as a traffic gap acceptance problem that focuses on identifying the gaps pedestrians feel comfortable crossing [15], [16]. A *traffic gap* is defined as the time taken by the closest vehicle to reach the pedestrian's longitudinal position. Yannis et al. [15] developed a logistic regression model to predict gap acceptance based on pedestrian waiting time, vehicle distance, age, and gender of pedestrians. More recent approaches use pedestrian's pose, motion, and vehicle behavior and develop Markovian [5] or Neural Network models [17], [18] to predict the crossing decision. These models identify if the pedestrian is standing or walking from their pose (represented by the joint positions of their skeleton model) and predict their future positions based on the continuous motion associated with the identified action. However, these models are unable to accurately predict changes in the actions, especially from standing to walking. Thus these models have limited application to crosswalk scenarios as they do not predict crossing decision(s) of pedestrians already waiting (and standing) at the crosswalk. Further, these models in addition to the pedestrian's position, require rich information, such as the pedestrian's pose which would require additional processing [17].

### B. Pedestrian Trajectory Prediction Models

Existing models for pedestrian trajectory prediction can be broadly categorized into three types – physics-based, planning-based, and trajectory-based models. Physics-based models express pedestrian motion either as individual kinematic models (constant velocity, acceleration or turn) or using an Interacting Multiple Model (IMM) framework combining the above individual kinematics [19], [20]. These models cannot reliably predict over 1 s in crossing scenarios as they are unable to predict motion changes such as turning at crosswalks. Recently, studies have modeled crossing decisions to improve the trajectory prediction [4], [5]. For example, Kooij et al. [4] modeled a pedestrian crossing laterally as a switched linear dynamical system. They used contextual cues such as vehicle-pedestrian distance, and head-orientation to identify if an approaching pedestrian will stop at the curb or continue to cross the road. However, these studies [4], [5] considered only laterally approaching pedestrians, but did not consider the behavior of pedestrians already waiting at the crosswalk and still had short prediction horizons ($\cong$ s).

Planning-based models represent pedestrian behavior as a Markov Decision Process, use pedestrian goal locations, and formulate the motion prediction as a optimal planning problem [21], [22]. These models, by attributing goal-seeking behavior to pedestrians, were able to do long-term predictions. Trajectory-based models are used to predict future pedestrian trajectories based on their past trajectories. These methods do not assume any pedestrian dynamics, but instead, learn the dynamics from the observed data. A common approach is to cluster the trajectories from the observed data using Gaussian process [23] or vector fields [24] and learn the motion patterns. More recently, deep learning models [8], [25] have been developed to predict pedestrian trajectories using observed trajectories. However, the planning-based and trajectory-based models are limited in their application to crosswalk scenarios, as they do not explicitly incorporate pedestrians' waiting behavior [20].

### C. Pedestrian-AV Interactions

Existing studies have compared pedestrian behavioral differences between AVs and HDVs. For example, Habibovic et al. [10] used a Wizard-of-Oz (WOZ) AV and found that pedestrians were less comfortable crossing in front of the AV than an HDV. On the contrary, Rothenbucher et al. [12] and Palmeiro et al. [11] used a WOZ AV and found that people crossed the street similarly in front of AVs and HDVs. Currently, there is no consensus on differences/similarities in pedestrian crossing behavior between AVs and HDVs, which warrants further investigation. Further, such studies [10]–[12] conducted in the real world have a limited range of vehicle behaviors, as the vehicle speeds are low for safety reasons.

## III. EXPERIMENTAL DATA COLLECTION

We developed an Immersive Virtual Environment (IVE) using Unity Game Engine (Unity Technologies, San Francisco, CA) to collect pedestrian-AV interaction data. The experimental setup (refer Fig. 2) consists of a virtual reality headset (Vive; HTC Corp., New Taipei, Taiwan), an eye tracker (Pupil, Pupil Labs, Berlin, Germany) and an omnidirectional treadmill (Omni;Virtuix Inc., Austin, TX).

We conducted a user study with 30 participants (mean age = 22.5 years, s.d. = 2.8 years). During the experiment, participants crossed a street at an unsignalized mid-block crosswalk with several oncoming fully automated vehicles.

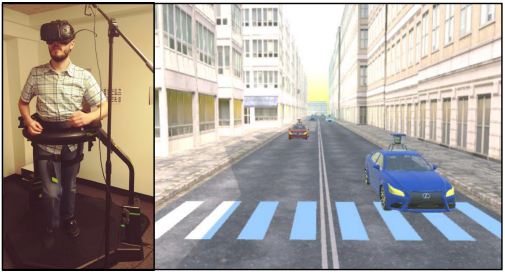

Fig. 2. Virtual Reality setup and the simulation environment for the user study. Pedestrians walk on the treadmill, which is translated to the virtual environment for improved immersion.

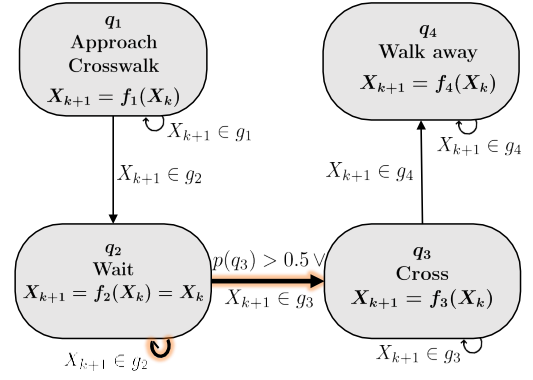

Fig. 3. Hybrid automaton of a rational pedestrian with the intent to cross. The bold arrows represent the transitions from the gap acceptance model.

The street was one-way with two lanes for the AVs. We manipulated the AV driving behavior (defensive, normal, or aggressive). We defined the driving behaviors as shown in Table I.

Participants were asked to perform a simple task of moving balls from one side of the road to the other, three times, which inherently made them cross the street six times per driving condition, resulting in a total of 540 crossings. Participants underwent a training session during which they got familiarized with the AVs' appearance. During the training session, AVs were travelling with a constant speed of 35 mph (15.6 m/s) and the participants always stayed on the sidewalk. During the actual treatment conditions, AVs spawned one after the other, 150 m away from the crosswalk, with a random time gap of either 3 s or 5 s between spawns and approached the crosswalk at a constant speed of 15.6 m/s (35 mph). For each treatment condition, the vehicle reacted according to its driving behavior when encountering a participant within its reaction distance (refer Table I). The different vehicle reactions (refer Table I) and spawn time gaps resulted in a varied range of AV speeds and spacing distances between consecutive AVs. Participants were encouraged to cross quickly and safely as their bonus payment was tied to the measures quantifying their performance (task completion time) and safety (distance from vehicle).

## IV. CROSSING BEHAVIOR HYBRID MODEL

We propose a hybrid automaton [26] to model pedestrian crossing behavior. The hybrid automaton has four discrete states—approach crosswalk, wait, cross, and walk away—each with an associated continuous motion. The sequence of these discrete states and the state transitions, together with the pedestrian trajectory, gives rise to the pedestrian's crossing behavior. The transition from wait to cross state is modeled as pedestrian's decision to cross, which is explained in section III-A. The proposed hybrid automaton model is shown in Fig. 3 and is formally defined as a tuple $(X, Q, f, G, T, R)$, where

- $X = (x, y, v_x, v_y) \in \mathbb{R}^4$, are the continuous pedestrian states – Cartesian positions $(x, y)$ and velocities $(v_x, v_y)$.
- $Q \in \{q_1, q_2, q_3, q_4\}$, are the discrete states of approach, wait, cross, or walkaway respectively.
- $f : Q \times X \to \mathbb{R}^4$, represents the continuous dynamics.
- $G \in \{g_1, g_2, g_3, g_4\}$ is a set of guard conditions, where
  - $g_1 = \{X \mid v_x \neq 0 \wedge sign(x)\, v_x < 0\}$
  - $g_2 = \{X \mid v_x = 0 \wedge v_y = 0\}$
  - $g_3 = \{X \mid v_y \neq 0\}$
  - $g_4 = \{X \mid v_x \neq 0 \wedge sign(x)\, v_x > 0\}$.
- $T$ is the discrete state transition function given as,

$$T(Q_{k+1} | Q_k) = \begin{cases} q_1 & \text{if } X_{k+1} \in g_1 \\ q_2 & \text{if } X_{k+1} \in g_2 \\ q_3 & \text{if } X_{k+1} \in g_3 \vee p(q_3) > 0.5 \\ q_4 & \text{if } X_{k+1} \in g_4 \end{cases}$$

  where $p(q_3)$ is the probability of crossing.
- $R : X \to X$ is a reset map of continuous velocity states after the discrete transitions.

We assume that pedestrians' velocity remains constant within each state and express the pedestrian dynamics using a constant velocity model with zero-mean Gaussian process noise $(W)$ as $X_{k+1} = f(X_k, \mathbf{0}, W)$. We also assume that pedestrians always enter the wait state before crossing. The wait is instantaneous when the pedestrian has decided to cross while in the approach state. The discrete state transitions are triggered when the corresponding guard conditions $(G)$ are satisfied. Additionally, the wait to cross transition can also be triggered by the decision to cross $(p(q_3) > 0.5)$.

TABLE I
PARAMETERS AND VEHICLE REACTIONS TO VARIOUS PEDESTRIAN POSITIONS FOR DIFFERENT DRIVING BEHAVIORS.

| Behavior | Pedestrian Position | | | | Reaction distance | Stopped distance | Maximum acceleration | Slow speed | Full speed |
| --- | --- | --- | --- | --- | --- | --- | --- | --- | --- |
| | Sidewalk | Wait Area | Same lane as AV | Other lane as AV | | | | | |
| Defensive | Full speed | Slow speed | Stop | Stop | 50 m | 3 m | 3 m/s$^2$ | 4 m/s | 15.6 m/s |
| Normal | Full speed | Slow speed | Stop | Slow speed | 30 m | 2 m | 5 m/s$^2$ | 7 m/s | 15.6 m/s |
| Aggressive | Full speed | Full speed | Stop | Full speed | 10 m | 1 m | 8 m/s$^2$ | NA | 15.6 m/s |

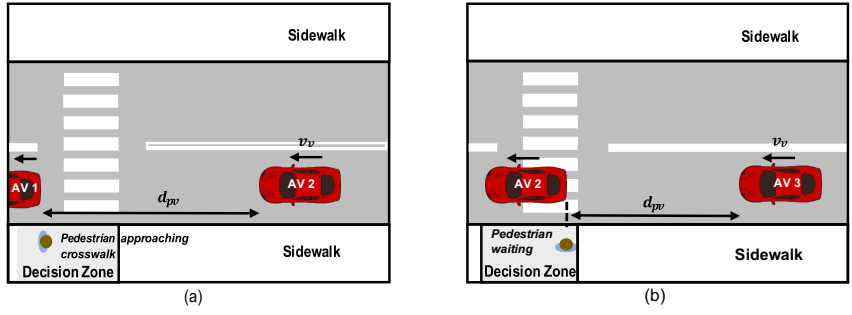

Fig. 4. Evaluation of gap acceptance: (a) a pedestrian is approaching and close to the crosswalk and a gap starts, and (b) a pedestrian is waiting on the road and a gap starts. Pedestrians' decide to accept/reject the gaps when they are in a decision zone, D.

## A. Crossing Decision Model

We describe pedestrian discrete state transition of wait to cross (i.e., crossing decision) through their gap acceptance behavior [15]. We develop a model that outputs the probability of accepting a traffic gap. Since gap acceptance is a discrete phenomenon, we assume the following.

- The decision to accept/reject a gap is made at the start of a gap, and the decision holds for the entire duration of that gap. A gap starts when a vehicle just passed the pedestrian (refer Fig. 4) and is considered to be accepted when the pedestrian starts crossing during that gap.
- Pedestrians always use the crosswalk for crossing the road and gaps are accepted only when the pedestrian is close to the crosswalk, denoted by the decision zones, D, in Fig. 4.
- Gaps are evaluated when the pedestrian is in the wait state or in the approach crosswalk state and within the decision zone. When a gap is accepted while the pedestrian is approaching, they enter the wait state for an infinitesimal time and transition to the cross state.

We model gap acceptance using a Support Vector Machine (SVM) classifier and obtain probabilistic outputs following the method in [27]. The input parameters to the SVM model were identified from literature [28], [29] as significantly affecting pedestrian crossing behavior and are detailed in Table II and are denoted by $F_i$, where $i$ is the time step.

## B. Hybrid Model usage for Real-Time Trajectory Prediction

We incorporate the constant velocity continuous dynamics within a Kalman filter framework and the SVM gap acceptance model in our hybrid automaton model for tracking pedestrians' position in real time. We tune the process and measurement noises to obtain the best tracking performance [30]. For evaluating the traffic gap, we define a set $D$ as the decision zone given by $D = \{X \mid |x| < 3\}$.

We evaluate the hybrid model by predicting pedestrian's trajectory during crossing. Algorithm 1 shows the steps for real-time pedestrian trajectory prediction. When the pedestrian is within the sensing range of the AV, the vehicle and pedestrian measurements are used to calculate the initial pedestrian states and the initial value of the features (refer Table II). The inference framework has two stages, predict and update. During the predict stage, the pedestrian motion is predicted using the continuous motion model and the discrete state is updated based on which guard condition the continuous state satisfies. Additionally, the transition from wait to cross can be predicted using the SVM gap acceptance model. The gap acceptance is evaluated when the pedestrian is within the decision zone and when a gap starts. The prediction loop continues for every time step within the prediction horizon $N$, using the previously predicted measurements. During the update stage, the states are updated based on the prediction for that time instant and the measurement.

TABLE II
INPUT PARAMETERS FOR GAP ACCEPTANCE MODEL.

| Parameter | Description |
|---|---|
| AV distance [m] | Longitudinal distance between AV and pedestrian |
| AV speed [m/s] | Speed of the AV |
| Wait time [s] | Time elapsed since pedestrian started waiting |
| Gaze ratio | Proportion of time pedestrian looked at AVs in the previous second |
| Curb distance [m] | Lateral distance between pedestrian & road edge |
| CW distance [m] | Longitudinal distance between pedestrian & crosswalk |
| Ped. speed [m/s] | Average pedestrian speed in the previous second |

---

**Algorithm 1** Inference framework for Hybrid model

1: Initialize $X_1, Q_1, F_1, i \leftarrow 1$
2: **while** $t_i < T$ **do**
3:    **procedure** PREDICT
4:    ▷ Predict discrete and continuous states for entire prediction horizon
5:    **for** $k \leftarrow 1$ to $N$ **do**
6:       **if** (Gap Starts $\wedge$ ($\hat{X}_{i+k} \in D$) $\wedge$ $\hat{Q}_{i+k} == q_2$) **then**   ▷ check gap acceptance probability
7:          Calculate $P(q_3) = P(\hat{Q}_{i+k+1} = q_3, \hat{F}_{i+k+1})$
8:       **end if**
9:       Sample $t_{cross}, v_{start}$ if $P(q_3) > 0.5$
10:       $\hat{Q}_{i+k+1} \leftarrow h(\hat{X}_{i+k}, P(q_3), t_{cross})$
11:       Reset $\hat{X}_{i+k}$ if $\hat{Q}_{i+k+1} = \hat{Q}_{i+k}$
12:       $\hat{X}_{i+k+1} \leftarrow f(\hat{X}_{i+k}, \hat{Q}_{i+k+1})$
13:       Evaluate $\hat{F}_{i+k+1}$
14:       $k \leftarrow k + 1$
15:    **end for**
16:    **end procedure**
17:    **procedure** UPDATE
18:       $X_{i+1} \leftarrow$ update state given $\hat{X}_{i+1}$, measurement $z_{i+1}$
19:       $Q_{i+1} \leftarrow h(X_{i+1})$
20:    **end procedure**
21:    $i \leftarrow i + 1$
22: **end while**

There is a delay between deciding to cross and start of crossing [31]. We express this delay as $t_{cross}$ and sample it from an exponential distribution learned from the collected data. The transition from wait to cross occurs when it is $t_{cross}$ seconds since the time gap was accepted (i.e., time delay is reached), expressed by the function $h$. Similarly, the pedestrian speed when starting to cross, $v_{start}$, is sampled from a Gaussian distribution learned from the data. During prediction stage, we assume constant velocity dynamics for the AVs. When no measurement is available, the features $F_i$ within the prediction horizon are calculated from the predicted pedestrian and AV positions. We assume that gaze ratio of pedestrians remains the same for the entire prediction horizon as the most recent observation. Using predicted data instead of actual measurements for the features, the hybrid model is able to perform long-term trajectory prediction.

## V. RESULTS

### A. Crossing Decision Prediction

We used the SVM gap acceptance model for predicting the crossing decision. For model training and testing, we extracted actual observed data for the instances when gaps are evaluated, i.e., when gaps start, that resulted in a total of 508 accepted gaps and 1195 rejected gaps. We split the data into training (80 %) and testing (20 %) sets. We used a cubic kernel for the SVM model. We compared the model with two baselines—a logistic regression model, similar to [15], and a model trained using the conditional probability distributions of gap acceptance conditioned on the observations mentioned in Table II. We used F1-score for model comparison. As shown in Table III, the SVM model performs better than the baselines and is used for pedestrian trajectory prediction. Table IV shows the importance of the various parameters based on the SVM model performance [32]. The features are arranged in descending order of importance. Gaze ratio has the least impact on the model performance.

### B. Hybrid Model Real-time Trajectory Prediction

The developed hybrid model was used for predicting pedestrian crossing trajectories. The transition from wait to cross was predicted using the SVM gap acceptance model. We used the following metrics to evaluate the trajectory prediction performance.

TABLE III
COMPARISON OF GAP ACCEPTANCE MODELS.

| Model | Accuracy | Precision | Recall | F1-Score |
|---|---|---|---|---|
| Probability Distributions | 0.76 | 0.66 | 0.51 | 0.56 |
| Logistic Regression | 0.79 | 0.71 | 0.52 | 0.60 |
| Support Vector Machine | 0.88 | 0.75 | 0.73 | 0.74 |

- Average displacement error (ADE): mean distance between predicted and actual trajectories for horizon $N$.
- Final displacement error (FDE): distance between predicted and actual position at last time step of $N$.
- Root mean squared error (RMSE) between the predicted and actual trajectories.

TABLE IV
SVM GAP ACCEPTANCE MODEL FEATURE RANKING.

| Feature removed | Accuracy | Precision | Recall | F1-Score |
|---|---|---|---|---|
| AV distance | 0.85 | 0.74 | 0.55 | 0.63 |
| CW distance | 0.86 | 0.71 | 0.66 | 0.69 |
| Curb distance | 0.86 | 0.71 | 0.71 | 0.71 |
| Wait time | 0.87 | 0.72 | 0.72 | 0.72 |
| AV speed | 0.88 | 0.78 | 0.68 | 0.73 |
| Ped. speed | 0.87 | 0.74 | 0.71 | 0.73 |
| Gaze ratio | 0.88 | 0.76 | 0.75 | 0.75 |

Current collision avoidance systems use a constant velocity model for pedestrian trajectory prediction [34]. Thus, similar to [4], we compared our hybrid model against a baseline constant velocity model without any discrete states. We report the trajectory prediction performance at varying prediction horizons. From Fig. 6, it can be seen that the hybrid model performs better than the constant velocity model across all metrics. Also, the difference in performance between the two models grows with the prediction horizon.

### C. Pedestrian Behavior Comparison

We compared two measures, gap acceptance and walking speed, describing crossing behavior during wait to cross transition and during approach and cross states respectively, with published results from real-world studies [15], [33]. For comparison validity, we chose real-world studies that had a similar road structure (two-lane uncontrolled mid-block crossing) as our IVE.

*1) Gap Acceptance:* We compared gap acceptance behavior in AV interactions with [15] and used the same gap measure as [15]. *Accepted traffic gaps* were the difference between two time points: the time when the pedestrian just stepped onto the road and the time when the head of the vehicle had just passed the pedestrian's longitudinal position. The comparison of the cumulative gap acceptance distributions is shown in Fig. 5 (a). We calculated KL-divergence [35] to compare the curves and find the value to be 0.17 (low value implies more similarity).

*2) Walking Speed:* We calculated pedestrian speed as the finite difference of their positions and applied a moving average filter to reduce noise. Pedestrians tend to walk faster while crossing ($1.58$ *m/s*) [33] than on sidewalks ($1.48$ *m/s*). We observe a similar trend in pedestrian speeds in AV interactions ($1.68$ *m/s* while crossing compared to $1.52$ *m/s* while on sidewalk), as shown in Fig. 5 (b).

## VI. DISCUSSION

The present study aimed at developing a model to characterize long-term pedestrian crossing behavior which in turn can be used for long-term pedestrian trajectory prediction. The hybrid systems model we developed incorporates pedestrian crossing decision-making and accounts for pedestrians already waiting at the crosswalk. This makes the model suitable for pedestrian behavior (trajectory) prediction at crosswalks. Trajectories predicted using our hybrid model had lower errors than the baseline model, and the performance difference increased with the prediction horizon. Thus our model is also better suited for longer prediction horizons than existing models [4], [5]. Further, our hybrid model includes the contextual information of the vehicle behavior (through vehicle distance and speed).

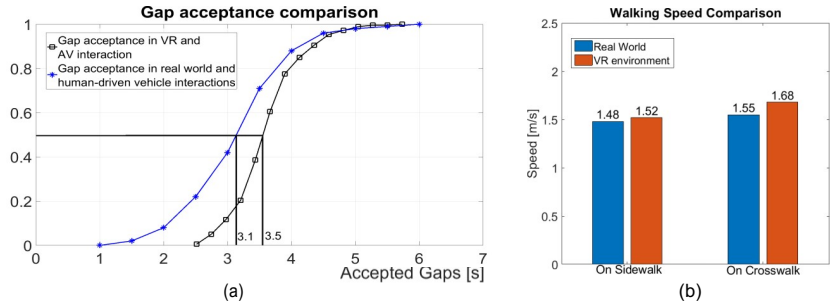

Fig. 5. Comparison of measures of crossing behavior. (a) Similar cumulative probability curves for gap acceptance in both AV and HDV scenarios [15]. (b) Higher walking speeds observed while crossing than on sidewalk, in both AV and HDV cases [33].

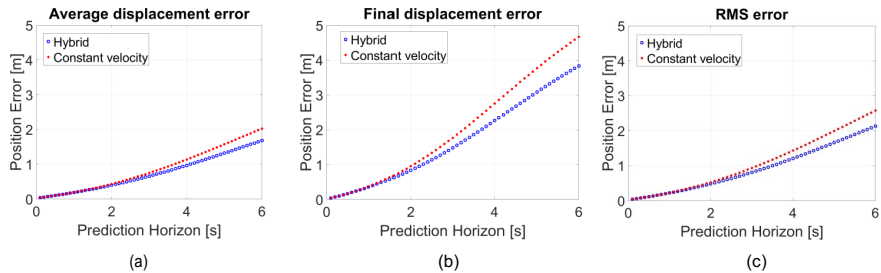

Fig. 6. Pedestrian tracking comparison for different prediction horizons for (a) Average displacement error, (b) Final displacement error, (c) RMS error. The hybrid model has lower error than the constant velocity baseline across all three metrics.

The SVM gap acceptance model we developed can function with only the pedestrian's position information, while many existing models [5], [17], additionally require rich pedestrian pose information as discussed in Section II.

We should acknowledge that unlike previous studies [36], [37], we did not find a substantial relationship between pedestrian crossing and gaze behaviors. Including gaze ratio did not substantially improve the gap acceptance predictions (refer Table IV). This could be because any rational pedestrian intending to cross can be expected to always look for the vehicles irrespective of their decision to cross or not.

In addition, we also examined and found similarities between measures of crossing behavior, namely, gap acceptance and walking speed, during interactions with AVs in IVE and HDVs in the real world. These results are in line with [11], [12], where pedestrians behaved normally around AVs as they would around HDVs. The similarities suggest the applicability of the hybrid model developed from IVE for real-world scenarios with both AVs and HDVs.

## VII. CONCLUSIONS AND FUTURE WORK

In this paper, we developed an accurate long-term pedestrian crosswalk behavior (trajectory) prediction model. The developed model can aid safe AV-pedestrian interactions, which can improve pedestrian acceptance of AVs. The similarities identified in crossing behavior between AV interactions in IVE and published results in real-world studies with HDV interactions demonstrated relevance of the developed model to real-world scenarios with both AVs and HDVs and the potential use of IVEs to study pedestrian-AV interactions.

In this work, we focused on individual pedestrians who always had the intent to cross. However, in the real world, multiple pedestrians are interacting with each other. Also, AVs would need to identify crossing from non-crossing pedestrians for safer interactions. Both of these could be considered for the future. Future work can focus on incorporating more contextual information such as pedestrian attributes, environmental factors, etc., in the discrete state transition model or the motion model or both. Though we maintained comparison validity by choosing studies with similar road and traffic conditions, we acknowledge there could still be potential behavioral differences. A future study could consider pedestrians interacting with HDVs in an IVE, although engaging multiple human drivers through the IVE may be a challenge. Also, future work could focus on validating the hybrid model on a real-world data set.

## VIII. ACKNOWLEDGMENT

Toyota Research Institute ("TRI") provided funds to assist the authors with their research but this article solely reflects the opinions and conclusions of its authors and not TRI or any other Toyota entity. This research was supported in part by by the Automotive Research Center (ARC) at the University of Michigan, with funding from government contract DoD-DoA W56HZV14-2-0001, through the U.S. Army Combat Capabilities Development Command (CCDC) /Ground Vehicle Systems Center (GVSC), and in part by the National Science Foundation.

DISTRIBUTION A. Approved for public release; distribution unlimited. OPSEC# 3814